# A Multi-Task Learning Approach for Meal Assessment

Ya Lu[1], Dario Allegra[2], Marios Anthimopoulos[1], Filippo Stanco[2], Giovanni Maria Farinella[2], Stavroula Mougiakakou[1, 3]

[1] ARTORG Center for Biomedical Engineering Research, University of Bern, Switzerland

[2] Department of Mathematics and Computer Science, University of Catania, Italy

[3] Department of Endocrinology, Diabetes and Clinical Nutrition, Bern University Hospital "Inselpital", Bern, Switzerland

{ya.lu, marios.anthimopoulos, stavroula.mougiakakou}@artorg.unibe.ch
{allegra, fstanco, gfarinella }@dmi.unict.it

## ABSTRACT

Key role in the prevention of diet-related chronic diseases plays the balanced nutrition together with a proper diet. The conventional dietary assessment methods are time-consuming, expensive and prone to errors. New technology-based methods that provide reliable and convenient dietary assessment, have emerged during the last decade. The advances in the field of computer vision permitted the use of meal image to assess the nutrient content usually through three steps: food segmentation, recognition and volume estimation. In this paper, we propose a use one RGB meal image as input to a multi-task learning based Convolutional Neural Network (CNN). The proposed approach achieved outstanding performance, while a comparison with state-of-the-art methods indicated that the proposed approach exhibits clear advantage in accuracy, along with a massive reduction of processing time.

## 1 INTRODUCTION

Meal assessment in terms of calories and macro-nutrient content estimation increasingly becomes more and more important for individuals that want to follow a healthy lifestyle. Traditionally, dietary assessment is based on self-maintained dietary records and food frequency questionnaires [1], which are time consuming, expensive and pruned to errors. In addition, studies have shown that even well-trained end-users, such as individuals with type 1 diabetes, cannot precisely estimate the meal's nutrient content [2]. Therefore, innovative approaches able to real-time, reliable and accurate estimate the nutrient content of meals are highly desirable.

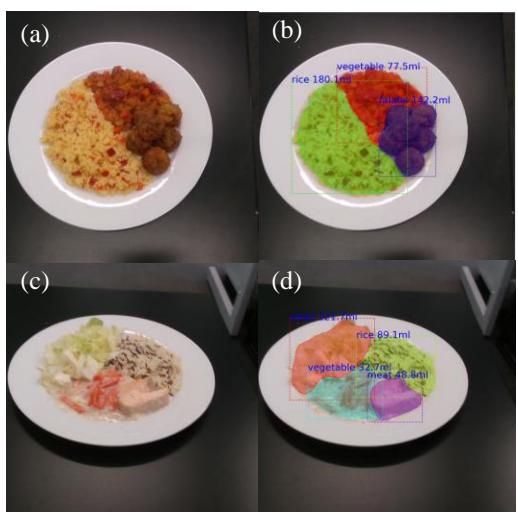

**Figure 1: Some results of the proposed method: (a) and (c) are the input images, while (b) and (d) are the output results**

The last years, research is conducted in calories and nutrients estimation by directly analyzing the meal's image as taken by the smartphone's camera. Nowadays, smartphones are equipped with high-quality camera and hardware that allows the execution of artificial intelligence and computer vision algorithms on the phone. In an ideal scenario, the users need to install a dedicatedly designed application in a mobile phone. By analyzing one or more meal images acquired by the phone's camera, the food type and the associated nutrient content could automatically be identified. The analysis usually involves four stages: i) food item detection and segmentation, ii) food recognition, iii) volume estimation and nutrient content calculation. Among them, the performance of the first three stages highly relies on the used AI algorithm and food multimedia dataset available, while the last on the quality of food composition database.

The first two stages are usually treated by directly employing well established segmentation and recognition methods [3-8], while food volume estimation tends to be addressed by geometric-based approaches involving more than one meal images and execution times longer than 5 seconds [9, 10].

In this paper, a multi-task learning (MTL) method is proposed aiming to realize the first three stages through one single network. MTL is a subfield of machine learning, which aims to improve the generalization performance of multiple tasks in parallel by sharing representations among these tasks [11]. In the proposed approach, an RGB food image is fed to the network; food segmentation and recognition are obtained by using instance segmentation-based approach, while food volume is inferred by the predicted food depth image and segmentation result.

Fig. 1 presents two examples of the proposed method, where (a) and (c) are single RGB image inputs, and (b) and (d) are outputs containing the pixel-wise segmentation map, as well as the corresponding food categories and respective volumes. The execution time of the entire procedure is 0.2 sec / image using GPUs. The performance, in terms of accuracy, of food segmentation and volume estimation outperforms the state-of-the-art.

## 2 RELATED WORK

### 2.1 Food Segmentation

Many of the well-known image segmentation algorithms have been applied to food images. At early stage, Shroff et al. [3] used adaptive thresholding method for food segmentation, which can only work under simplified condition where the plate must be white, and all the food items clearly spatially separated in. Anthimopoulos et al. [4] employed mean-shift clustering in the CIELab color space to achieve multiple connected food segmentation in a given plate. Dehais et al. [5] used traditional region growing assisted by CNN based border detection, achieving better performance, but in cost of computational resources. Aguilar et al. [6] combined food/nonfood binary mask and food localization bounding box to get pixel level food segmentation map, which is however invalid when there is large overlap among food localization bounding boxes.

### 2.2 Food Recognition

The food recognition problem is strongly related to the availability of food image databases. The quality and size of public food databases determine the performance of food recognition algorithms, therefore building a reliable database is of critical importance. The first published database was Pittsburgh Fast Food Images Dataset (PFID) [12], which contains 101 food items from 11 fast food chains. After that, a dataset named "Food101" with larger size was presented [13], containing 101,000 images belonging to 101 food classes. More recently, Matsuda et al. [14] proposed the UEC FOOD 100 database, in which 9060 food images referring to 100 Japanese food types are dedicatedly built. As an updated version with respect to UEC FOOD 100, UEC FOOD 256 [15] extended the food classes and number of images to 256 and 31651, respectively. To further promote the development of food recognition, a database involving 211 fine-grained food categories with 101733 training images, 10323 validation images and 24088 testing images is published, which has been employed for on-going iFOOD2018 food recognition challenge [16].

From algorithmic point of view, Shroff, et al. [3] proposed the first food recognition system using color, texture, and shape features for four kinds of food. Anthimopoulos et al. [17] proposed a Bag-of-Feature (BoF) model-based method for automatic diabetic food recognition. Farinella et al. [18] proposed an "Anti-Texons" feature representation approach to further enhance the accuracy of food recognition. Recently, deep CNN have been used for food recognition [7, 8, 12], which significantly improved the accuracy on large food databases. An accuracy of 89% has been achieved on Food101 [13] and 83.15% on UECFood256 [19]. These performances are obtained using fused multiple neural network architectures and deep residual network, respectively.

### 2.3 Food Volume Estimation

The first food volume estimation system was proposed by Chen, et al. [20]. It used a single view image as input, required a dedicated shape model for each food category and a calibrated reference card. Puri et al. [9] used a dense multi-view 3D reconstruction approach, which generated the 3D point cloud of the food based on a video sequence and plate-sized reference patterns. Recently, Dehais et al. [10] proposed a two-view 3D reconstruction approach using a credit card sized reference card. The approach was extensively tested on real dishes of known volume, and achieved an average error of less than 10% in 5.5 seconds per dish. The methods are integrated into a smartphone application, named GoCARB, and has been validated within a clinical trial, showing that the method is able to estimate the carbohydrate content on meals at free-conditions and indicating that the use of such an application positively impact the glucose control of individuals with diabetes [21].

### 2.4 Depth Prediction

Inferring depth image from RGB images has been extensively researched over the last years [22-25]. However, in the field of dietary assessment only a limited number of work has been reported mainly due to lack of data. A first attempt has been reported by Allegra et al. [26], which applied a SegNet-based [27] CNN architecture for single food image depth prediction. Recently, Christ et al. [28] proposed a CNN architecture with skip connections for food image depth prediction.

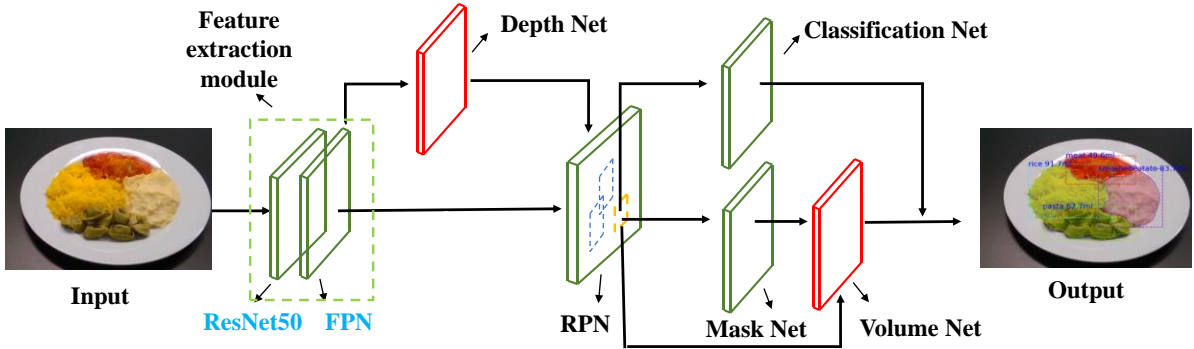

Figure 2: Multi-task learning network architecture

# 3 METHOD

## 3.1 Network Architecture

The introduced method is based on instance segmentation extended to additionally perform volume estimation. Instance segmentation performs segmentation and recognition simultaneously, using both semantic segmentation and instance identity [29, 30]. Driven by the effectiveness of region-based convolutional neural network (R-CNN) approach [31], He et al. [32] proposed the Mask R-CNN framework for instance segmentation, which applies an additional branch based on Faster R-CNN framework [33]. This Mask R-CNN approach surpasses all previous state-of-art methods for instance segmentation task [32]. Instance segmentation has proven to perform well for the object occlusion or near-connection cases, implying a potential use for food segmentation and recognition.

Differently than in the Mask R-CNN classic architecture, we extend it with newly designed modules conducting volume estimation. The overall network architecture of the proposed algorithm is presented in Fig. 2. It contains six main components: feature extraction module, depth prediction net, Region Proposal Net (RPN), recognition net, mask prediction net, and volume estimation net.

Among them the modules in green are maintained from Mask R-CNN [32], being responsible for food recognition and segmentation based on the 2D information of the image. In particular, the feature extraction module is composed by ResNet50 [34] and Feature Pyramid Network (FPN) [35]. The working principle basically consists of two stages: at the first stage, the RPN along with the feature extraction module preliminary produce the bounding box for each candidate object; at the second stage, the features extracted before, are identically resized associated to all the nominated bounding boxes, based on which the recognition, bounding-box regression and binary mask prediction are executed.

The red modules illustrated in Fig. 2 represent the new components added in the network for food volume estimation. Different from the previously introduced modules, the output from the "Volume Net", which reports the final volume value, requires 3D information of the food objects. In other words, the 3D features of the image must be extracted in advance, which is realized by adding a depth prediction branch implemented between the feature extraction and RPN modules. In this module, a depth map is used as ground truth for performing supervised learning. Note that since the final volume estimation module uses the resized 2D and 3D features, the scale information of each food candidate, which is essential for calculating volume, is not able to be maintained. In order to tackle this issue, we propose to convert the predicted depth into point cloud, which feeds to the RPN module. The detailed description of the newly introduced "Depth Net" and "Volume Net" are presented in the following.

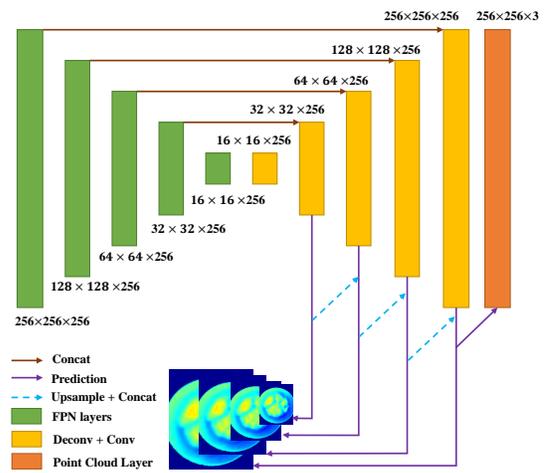

Figure 3: Architecture of depth net. Green layers are encoder part provided by FPN, while yellow layers are decoder layer. Orange layer is the point cloud convention layer

**Depth Net:** The architecture of Depth Net is shown in Fig. 3, and is mainly based on an encoder-decoder design architecture with skip connections and multi-scale side predictions, followed by a point cloud layer. Since the encoder part requires pyramid feature layers from a single scale input, which are in total similarity with the FPN output in the previous feature extraction module, the latter (green layers in Fig. 3) is adopted to act as the encoder part of our Depth Net. The kernel size and output channel for all the convolutional/deconvolutional layers are 3 and 256, respectively, in order to adapt the FPN output. In the network presented in Fig. 3, all the convolutional/deconvolutional layers (in yellow) are followed by batch normalization layers and RELU activation, and the prediction layers (in purple) are followed sigmoid activation to constrain the prediction depth in the range of 0~1m. With this network, the depth image is predicted in 4 different scales in total, which range from 64×64 to 256×256. Among them the first three (3) smallest scales contribute to improve the depth prediction accuracy, and only the largest one is used for point cloud converting.

For a given largest depth map $d_I$, the 3D point cloud, denoted as $X_I^i = (x_I^i, y_I^i, z_I^i)$, $i \in 1, \ldots, w \times h$ ($w$ and $h$ are the image width and height, respectively), can be readily calculated according to the pinhole camera model:

$$X_I^i = \begin{bmatrix} x_I^i \\ y_I^i \\ z_I^i \end{bmatrix} = K^{-1} \begin{bmatrix} u_I^i \\ v_I^i \\ d_I^i \end{bmatrix} \quad (1)$$

where $(u_I^i, v_I^i)$ is the pixel position on image, $d_I^i$ is the depth value of the $i^{th}$ pixel, and $K$ is the camera matrix, which is represented as:

$$K = \begin{bmatrix} f_x & 0 & c_x \\ 0 & f_y & c_y \\ 0 & 0 & 1 \end{bmatrix} \quad (2)$$

where $(c_x, c_y, f_x, f_y)$ index the camera intrinsic parameters.

**Volume Net**: It is implemented to achieve the volume estimation of food objects based on the foreground mask of the food candidate from mask net and the resized features provided by RPN net. We apply two (2) convolutional layers with kernel sizes of 7 and 1, respectively, and the output channel for both convolutional layers is 1024. Both the convolutional layers in this module are followed by batch normalization and RELU activation function. In addition, 1 average pooling layer with kernel size of 2 is applied before the 2nd convolutional layer, and 1 dense layer is utilized as the last layer to get the regressed volume.

### 3.2 Loss Function

Formally, during training, we define a multi-task loss on each input image as:

$$L = \sum_S L_{depth}^S + \sum_{RoI}(L_{cls} + L_{bbox} + L_{mask} + L_{vol}) \quad (3)$$

where $S$ indexes over the 4 predicted scales for depth image, $RoI$ represents the food candidates output from RPN. In (3), the definition of recognition loss $L_{cls}$, bounding box loss $L_{bbox}$ and object mask loss $L_{mask}$ are identical as those in [32], while the $L_{depth}$ is defined as multi-scale mean absolute error loss. The $L_{vol}$ is the combination of absolute percentage loss and absolute loss:

$$L_{vol} = \left| \frac{\hat{v} - v^*}{v^*} \right| + \alpha \left| \hat{v} - v^* \right| \quad (4)$$

where $v^*$ is the groundtruth volume, $\hat{v}$ is the predicted volume, and $\alpha$ is the weighting for the absolute loss.

### 3.3 Implementation Details

The experiments in this paper are conducted on a machine with NVIDIA GeForce Titan X GPU under a Linux OS. The whole framework is implemented using the Keras [36] with Tensorflow backend [37].

We set most of the hyper-parameters following [32], which we found robust. As the concept in [32], a RoI is considered positive if it has Intersection of Union (IoU) with a ground truth bounding box lager than a threshold, and negative otherwise. Instead of using 0.5 as the threshold defined in [32], we set it to 0.7 targeting at achieving more precise volume value. Same with the use of the mask loss in [32], the volume loss is also valid only on positive RoIs. The threshold we set for positive RoI in the experiment is 0.7. The weight $\alpha$ in volume loss calculation of (4) is set as 0.01. The weight of ResNet50 is initialized with ImageNet pretrained weight. The camera matrix is set as default values of [1, 0, 0.5; 0, 1, 0.5; 0, 0, 1] for simplicity.

During training, all the input RGB-D image pairs are cropped into $1024 \times 1024$, and then the depth images are resized into 256×256. We trained the network for 60K iterations in total, with initial learning rate of 1e-3, which is set to 1e-4 after 40K iterations. We use SGD optimizer with weight decay of 1e-4 and a momentum of 0.9. To increase the image variability, we augment the dataset by considering left-right and up-down flips during training.

Testing procedure solely requires one single RGB image as input. The proposal number of bounding boxes for FPN is set as 1000. We run the RPN branch on these proposals to get 50 proposal candidates with the highest scores. Based on these, the recognition branch, mask branch and volume estimation branch are then applied.

## 4 EXPERIMENTAL RESULTS

### 4.1 Datasets

We used the Madima17 dataset [26] as the training and evaluation datasets, providing 80 central-European meals and 21 detailed food categories. Each meal contains 2-4 food items, all of them of known volume (ground truth). It must be mentioned that the 21 food categories are merged into 6 broad categories (i.e. potato, meat, carrot, pasta, vegetable, rice), in order to assure a sufficient number of images for each food category. For each meal 6 RGB-D image pairs at a fixed resolution of 1920×1080 are available, 4 captured at distances of 40cm and 60cm, with 90º and 60º angles of view, respectively, and the rest 2 from random position (distance and angle).

In the dataset, we allocate 60 meals for training, 10 meals for validation, and the remaining 10 meals for testing. To perform comparisons thoroughly, we build three types of testing datasets: 1) "fixed set", which contains the samples captured only with 90º angle at 40cm distance; 2) "free set", which only contains the 2 randomly chosen samples; 3) "full set", which contains all the 6 capture samples for each meal.

Note that despite the testing data is divided in three types, the training data exploits the full dataset, i.e., four (4) image pairs taken from fixed positions and two (2) image pairs taken from randomly positions, to keep the generality of the work.

## 4.2 Food Segmentation and Recognition

In this section, we firstly use region-based metrics [26, 38] and the confusion matrix to evaluate the performance of food segmentation and recognition, respectively, and then we simultaneously examine the performance of both segmentation and recognition by employing standard Average Precision (AP) measures [39].

For the region-based metrics, $S = \{S_i\}_{i=1}^{m}$ and $T = \{T_i\}_{i=1}^{n}$ are defined as the predicted and ground truth segments, respectively, where $m$ and $n$ are the number of segments in $S$ and $T$. Then, the two normalized directional indices representing the worst (i.e., the predicted segment has minimum overlap with the ground truth) and the average segmentation performance are given as

$$NI_{min}(T \rightarrow S) = Min_i\left(\frac{Max_j(|S_i \cap T_j|)}{|S_i|}\right) \quad (5)$$

$$NI_{sum}(T \rightarrow S) = \frac{\sum_i Max_j(|S_i \cap T_j|)}{\sum_i |S_i|} \quad (6)$$

For final evaluation, two reversed directions of each index are combined as:

$$F_x = \frac{2 \times NI_x(T \rightarrow S) \times NI_x(S \rightarrow T)}{NI_x(T \rightarrow S) + NI_x(S \rightarrow T)}, x = min \ or \ sum \quad (7)$$

Table 1 shows the result based on (7), for the food segmentation evaluation in cases of the proposed method, as well as the state-of-the-art, such as CNN based border detection method [26] and region growing/merging method [38]. Note that both latter ones require an additional step to remove the background from true dish segmentation in the image. Table 1 indicates that the proposed method outperforms the other two in terms of both $F_{min}$ and $F_{sum}$ in the cases of both 'fixed set' and 'full set', validating the advantage of the proposed method.

**Table 1: Comparison of segmentation method**

|  | Fixed set | | Full set | |
| --- | --- | --- | --- | --- |
| Method | $F_{sum}$(%) | $F_{min}$(%) | $F_{sum}$(%) | $F_{min}$(%) |
| Proposed | **94.36** | **83.90** | **94.10** | **78.18** |
| Method in [26] | 93.69 | 74.26 | - | - |
| Method in [38] | 92.47* | 73.36* | 91.83* | 75.33* |

*means the value is from our re-implementation

Fig 4 shows the confusion matrix of the proposed algorithm on the full set. It can be observed that "carrot", "pasta" and "vegetable" are perfectly recognized, while "potato" has the worst performance since 25% of it is mis-recognized with "meat". This is mainly due to the fact that some food made by potato, e.g., rösti (type of Swiss dish consisting of potatoes in the style of a fritter), looks very similar with a piece of roasted chicken breast. Furthermore, "rice" was often misclassified as "pasta", since in many cases both of them were served with sauce (and they had almost identical visual appearance). The average recognition rate over the six categories is 93.3%, indicating a good recognition performance of the proposed algorithm.

Table 2 indicates the overall performance of the proposed module following AP metrics, for both food segmentation and recognition, using 3 typical parameters namely $AP_{50}$, $AP_{75}$ and mAP [39]. Among them, $AP_{50}$ and $AP_{75}$ represent the percentage of the samples having the IoU value (between predicted segment and ground truth) larger than 0.5 and 0.75, respectively. and mAP is the average percentage of the samples with IoU thresholds from 0.5 to 0.95, which is expressed as:

$$mAP = \frac{1}{10}\sum_{IoU} AP_{IoU}, \ IoU \in [0.5:0.05:0.95] \quad (8)$$

|  | potato | meat | carrot | pasta | vegetable | rice |
| --- | --- | --- | --- | --- | --- | --- |
| potato | 75 | 25 | | | | |
| meat | | 91 | | 3 | | 6 |
| carrot | | | 100 | | | |
| pasta | | | | 100 | | |
| vegetable | | | | | 100 | |
| rice | | | | 16 | | 84 |

**Figure 4. Confusion matrix on the full set. The entry in the *i*th row and *j*th column corresponds to the percentage of images from class *i* that classified as class *j*.**

From the AP values one can observe that the performance of our algorithm on all the three datasets are similar, indicating that the module works well regardless of the image angle of view and distance. These metrics also set the baseline for the food segmentation and recognition tasks on this dataset.

**Table 2: Quantitative results using AP measures**

| Dataset | mAP (%) | $AP_{50}$ (%) | $AP_{75}$ (%) |
| --- | --- | --- | --- |
| Fixed | 69.4 | 90.4 | 85.7 |
| Free | 63.2 | 83.7 | 79.6 |
| Full | 64.7 | 85.1 | 79.1 |

## 4.2 Depth Image Prediction

In this section, we evaluate the performance of the newly introduced depth net module. Fig. 5 illustrates three examples of the input image, the ground truth depth map and the depth map predicted by our algorithm. Good matches between the ground truth and the predicted depth maps can be observed, demonstrating the viability of attaining depth image only from a single color input.

The performance of the proposed depth net module is further evaluated based on both Mean Absolute Distance (MAD) and Absolute Relative Difference (ARD) metrics, with quantified comparison to the state-of-the-art which uses the same dataset [26]. In such a comparison, only the pixels inside the plate are evaluated.

The results are reported in Table 3, showing that the proposed approach completely outperforms the conventional method in terms of both MAD and ARD, with the cases of both 'free set' and 'full set'.

## 4.3 Volume Estimation

In this section, the performance of volume estimation module is evaluated and compared to a conventional approach based on 3D reconstruction [10, 26]. The evaluation metric we use is average percentage error for each food item.

Due to the intrinsically distinct mechanisms between our method and the 3D reconstruction, the experimental conditions must be set for each separately. Whilst the proposed module requires only one RGB image as input, the 3D reconstruction method demands at least two images taken from different angle of views [10, 26]. In addition, in our MTL approach, the volume estimation relies on the predicted segmentation and depth, meaning that the performance might be degraded by the quality of other modules in the network. Nevertheless, the ground-truth segmentation map is set as the input of 3D reconstruction method in the experiment.

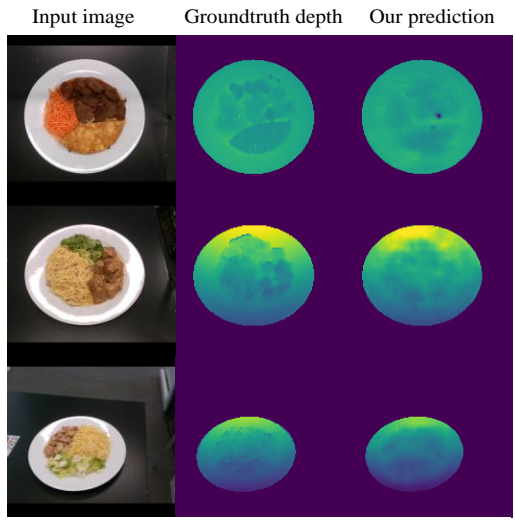

**Figure 5: Depth prediction result on Madima17 dataset**

**Table 3: Comparison of depth prediction method**

|         | Free set  |         | Full set  |         |
|---------|-----------|---------|-----------|---------|
| Method  | MAD (mm)  | ARD (%) | MAD (mm)  | ARD (%) |
| Proposed | **6.75** | **1.25** | **5.71** | **1.13** |
| Method in [26] | 8.64 | 1.76 | 6.03 | 1.25 |

**Table 4: Comparison of volume estimation**

|         | Food item's average percentage error | | | |
|---------|-------|-------|-------|----------|
| Method  | Fixed (%) | Free (%) | Full (%) | Process time (s) |
| Proposed | **17.5** | **19.1** | **19.0** | **<0.2** |
| 3D Reconstruction | 22.6 | 36.1 | 33.1 | 5.5 |

Table 4 reports the quantitative comparison on the three datasets described in Section 4.1. It can be observed that the 3D reconstruction only performs good with 'fixed set', implying that the method has high demand of the image angle of view. While the proposed module possesses smallest error in all 3 case (even with 'fixed dataset'), validating its robustness and high accuracy. More importantly, the process time of the proposed method is less than 0.2 s, being 25 times shorter than that taken by conventional 3D reconstruction. It has to be noted that in [26], the 3D reconstruction approach achieves a volume estimation error in the order of 13.8% on the fixed set. The difference is mainly due to the usage of a different evaluation metric. The volume estimation error in [26] is for each meal, while in the current research, due to the intrinsic mechanism of our approach, the volume error of each food item in the meal is calculated. The currently used evaluation metric is somehow more strict than the one used in [26].

## 4 CONCLUSIONS

In this paper, we have presented a MTL-based CNN for meal assessment, which simultaneously addresses food segmentation, recognition and volume estimation. The method achieved superior performance compared with state-of-the-art methods on the Madima17. The proposed method, by using only one RGB image as input, achieved: 1) an improved food segmentation - performance, in terms of $F_{min}$, has been significantly increased by 9% , 2) 3D information of the RGB food image is extracted by newly designed Depth and Volume nets, achieving more stable and accurate result comparing with the conventional approach based on 3D reconstruction; and 3) the computational time of the entire pipeline is 0.2 s - two orders faster than that for conventional volume estimation (5 s). Future work includes the extension of the methods to images with multiple dishes and databases with food higher diversities in terms of food categories and images per category.

## ACKNOWLEDGMENTS

This work is partially supported by the European Commission Seventh Framework Programme (FP7-PEOPLE-2011-IAPP) under Grant 286408.